# Semi-Supervised Facial Expression Recognition based on Dynamic Threshold and Negative Learning

Zhongpeng Cai, Jun Yu*, Wei Xu, Tianyu Liu, Jianqing Sun, Jiaen Liang

*Abstract*—Facial expression recognition is a key task in human-computer interaction and affective computing. However, acquiring a large amount of labeled facial expression data is often costly. Therefore, it is particularly important to design a semi-supervised facial expression recognition algorithm that makes full use of both labeled and unlabeled data. In this paper, we propose a semi-supervised facial expression recognition algorithm based on Dynamic Threshold Adjustment (DTA) and Selective Negative Learning (SNL). Initially, we designed strategies for local attention enhancement and random dropout of feature maps during feature extraction, which strengthen the representation of local features while ensuring the model does not overfit to any specific local area. Furthermore, this study introduces a dynamic thresholding method to adapt to the requirements of the semi-supervised learning framework for facial expression recognition tasks, and through a selective negative learning strategy, it fully utilizes unlabeled samples with low confidence by mining useful expression information from complementary labels, achieving impressive results. We have achieved state-of-the-art performance on the RAF-DB and AffectNet datasets. Our method surpasses fully supervised methods even without using the entire dataset, which proves the effectiveness of our approach.

## I. INTRODUCTION

Facial Expression Recognition (FER) aims to identify individuals' emotional states by analyzing facial images or videos. It is a broad research field that spans machine learning, image processing, psychology, and more, with a wide range of applications including safe driving, intelligent surveillance, and human-computer interaction. The task of expression recognition is a classic problem in the field of pattern recognition, typically involving the recognition of six basic emotions: happiness, surprise, sadness, anger, disgust, and fear. A schematic diagram of the 6 basic expressions is shown in the figure 1.

In the early days of the rise of deep learning, fully supervised methods [1]–[4] made significant progress in the field of expression recognition, and the introduction of multiple benchmark datasets (such as RAF-DB [5] and AffectNet

J. Yu, Z. Cai are with the Department of Automation, University of Science and Technology of China, Hefei, 230027, CN (e-mail: harryjun@ustc.edu.cn; zpcai@mail.ustc.edu.cn)

Corresponding author: J. Yu

W. Xu is with the First Affiliated Hospital, Division of Life Sciences and Medicine, University of Science and Technology of China, Hefei, China (e-mail: xw199807@163.com)

T. Liu is with the Jianghuai Advance Technology Center, Hefei, China (e-mail: liutianyu18@mails.ucas.ac.cn)

J. Sun and J. Liang are with Unisound AI Technology Co., Ltd, Shanghai, China (e-mail: sunjianqing@unisound.com; liangjiaen@unisound.com)

Manuscript received June 30, 2024

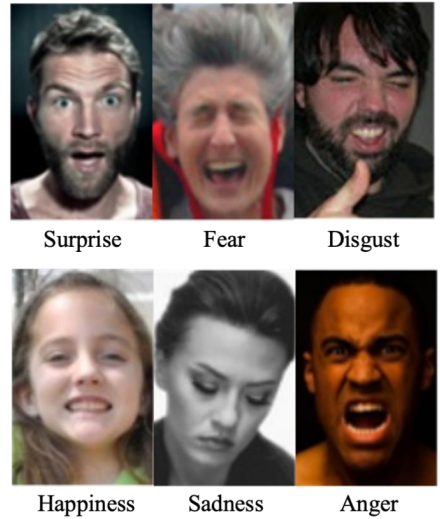

Fig. 1. The mainstream research in facial expression recognition primarily focuses on six categories, known as basic expressions, which are Happiness, Surprise, Sadness, Anger, Disgust and Fear.

[6]) also led to rapid development of expression recognition technology. However, in traditional fully supervised facial expression recognition methods, the prediction accuracy of the model heavily depends on a large amount of high-quality annotated data. Current datasets still have two major problems: (1) The scale of facial expression datasets is limited. Facial-related datasets are privacy-sensitive, difficult to collect, and expensive to annotate, requiring experts to do so. (2) There is an issue of class imbalance in expression recognition datasets. For example, in the AffectNet dataset, there is a huge disparity of up to ten times in the number of "happiness" samples compared to the "disgust" category.

For the field of expression recognition, the amount of data has always been a huge bottleneck. Compared to other recognition task datasets like ImageNet [7], which have tens of millions of samples, the scale of expression recognition datasets is still very limited. The limited scale of data leads to a limitation in model capacity, affecting the further development of the field of expression recognition.

The essence of deep learning models is to fit the data distribution of training data with a huge number of parameters. The problem of unbalanced data distribution in expression recognition leads to significant biases in deep learning models during training. The trained models may tend to predict categories with more samples, resulting in a lower recall rate for minority classes. This is not very friendly for the task of expression recognition.

To address the above issues, researchers have begun to study methods that use as little labeled data as possible or make the most of unlabeled data. Semi-supervised learning methods have started to be applied to the field of expression recognition.

We have adopted a semi-supervised approach, using uniform sampling of the number of samples for each emotional category to reduce model bias caused by imbalanced sample distribution. At the same time, semi-supervised methods can make full use of unlabeled data, providing possibilities for dataset expansion. Our designed framework can assign pseudo-labels to unlabeled data, helping the model learn facial expression features. Considering the varying difficulty of recognizing each category, we set a confidence threshold for each emotional category and gradually adjust this threshold as training progresses to make full use of unlabeled data. To maximize the value of unlabeled data, we specifically designed a selective negative learning module aimed at fully mining the information in those samples with low confidence by extracting beneficial information for expression recognition from complementary labels. The negative learning task is also friendly to noisy tasks, effectively mitigating the noise problem caused by pseudo-labels. In addition, we introduced a negative sample learning loss function to constrain the model to fully learn expression-related features from negative samples.

The contributions can be summarized as follows:

- We designed a feature extraction network for expression recognition to enhance local feature representation while using a feature map dropout strategy to ensure the model does not overfit to a particular local area. This provides more discriminative features for expression recognition.
- We introduced a Dynamic Threshold Adjustment (DTA) module, which sets confidence thresholds for each emotional category at different training stages to control the quality of generated pseudo-labels. This strategy allows the model to flexibly adjust the use of unlabeled data during training and better adapt to the dynamic changes in data.
- To make full use of unlabeled data, we also designed a Selective Negative Learning (SNL) module to fully utilize those samples with low confidence, mining useful expression information from complementary labels. We also introduced a negative learning loss to further enhance the model's performance.

## II. RELATED WORK

### A. Facial Expression Recogniton

Traditional methods in facial expression recognition have traditionally relied on handcrafted feature extraction for pattern recognition. Approaches include those based on texture features [8]–[12], geometric features [13], and hybrid feature methods [14]–[17]. While effective in extracting facial features, these methods often suffer from high computational complexity, limited accuracy, susceptibility to noise, and poor scalability. The advent of deep learning in computer vision has revolutionized facial expression recognition, with recent years witnessing substantial advancements in methods utilizing deep learning [1], [2], [4], [18]–[21]. Yang et al. [1] proposed De-expression Residue Learning (DeRL), a method leveraging conditional Generative Adversarial Networks (cGAN) to capture facial expression features by de-expressing faces. Their approach effectively preserved expression-related information in intermediate layers, achieving notable performance. With the rise of transformers in computer vision [22], transformer-based approaches have entered the field of facial expression recognition for feature extraction. However, initial attempts faced challenges due to convergence issues and a tendency to focus on occluded and noisy regions. TransFER [3] represents a notable transformer-based facial expression recognition method, utilizing multi-branch attention descent to address convergence challenges, albeit at the expense of increased computational demands. To mitigate these issues, Xue et al. [4] introduced two attention pooling modules, focusing on patch and token attention pooling. These modules guide the model to emphasize discriminative features while suppressing noise, without the need for learning additional parameters, resulting in an intuitive reduction in computational costs.

### B. Semi-Supervised Learning

With the widespread adoption of residual learning [23] and visual transformers [22] in deep learning, the parameterization of models has seen a significant increase. However, as the number of parameters grows, data scale becomes a critical limiting factor for the capabilities of deep learning models. In the domain of facial expression recognition, the high difficulty and cost associated with manual annotation further exacerbate the challenge. Consequently, leveraging unlabeled data has become a pressing issue in facial expression recognition. One crucial direction in utilizing unlabeled data is through semi-supervised learning. A popular category of semi-supervised learning methods involves generating pseudo-labels for unlabeled images and training models to predict these artificial labels [24], [25]. Similarly, consistency regularization methods [26]–[28] leverage the model's prediction distribution to obtain pseudo-labels through random modifications of inputs or model functions. Fixmatch [29] combines the strengths of both approaches, implementing consistent regularization through weak and strong data augmentations, obtaining pseudo-labeled data for samples with confidence levels exceeding a threshold. However, a limitation of this method is the fixed threshold, restricting its modeling capacity early in the training process. To address this challenge, Flexmatch [30] introduces the Curriculum Pseudo Labeling (CPL) method, a staged learning approach that adapts thresholds for different classes based on the model's learning state. This flexible adjustment enables the model to effectively utilize information-rich unlabeled data and their pseudo-labels at each time step. Additionally, Dash [31] selects samples by retaining only those with losses below a given threshold during each update iteration and dynamically adjusts the threshold through iterative refinement.

## III. OVERALL APPROACH

In this section, we introduce a basic framework for our semi-supervised framework. To better understand the semi-supervised framework, we first briefly introduce the fully supervised framework and then transition to the semi-supervised framework.

### A. Problem Formulation

We begin by introducing the fully supervised framework. For a fully supervised facial expression recognition task, there is a labeled dataset $D = \{(x_i, y_i)\}_{i=1}^N$, where $x_i$ represents the training image, $y_i$ represents the corresponding label category of the training image, and $N$ represents the number of samples. The conventional approach is to train a neural network model to predict the probability of each expression category, and to optimize the neural network using the cross-entropy loss constraint:

$$L_{CE}^s = -\sum_{c=1}^{C} y_i^c \log(p_c(x_i, \theta)),\qquad(1)$$

where $p_c(x_i, \theta)$ denotes the predicted probability that data $x_i$ belongs to category $c$ when the model parameters are $\theta$, $y_i^c$ is a binary indicator that is 1 if sample $x_i$ belongs to category $c$ and 0 otherwise, and $C$ is the total number of categories.

However, for semi-supervised facial expression recognition tasks, only a portion of the data is labeled, while the rest is unlabeled. In this case, the cross-entropy loss cannot be computed.

We divide the original training data into two sets, including a labeled set and an unlabeled set. Let $D_l$ be the labeled set:

$$D_l = \{(x_l, y_l)\}, l = 1, \ldots, N\qquad(2)$$

where $x_l$ represents the data sample, $y_l$ represents the corresponding label, and $N$ represents the number of labeled samples. The unlabeled data contains the same categories, denoted as $D_u$:

$$D_u = \{x_u\}, u = 1, \ldots, M\qquad(3)$$

where $M$ is the number of unlabeled training data.

Existing semi-supervised methods train a model using the labeled data, and then use this trained model to generate pseudo-labels $\hat{y}_u$ for the samples $x_u$. With these pseudo-labels, the originally unlabeled data can now be optimized together with the labeled dataset using the cross-entropy loss.

### B. Semi-Supervised Framework

As shown in Figure 2, our semi-supervised framework mainly consists of two parts: one for learning with labeled data and the other for learning with unlabeled data.

For labeled data $D_l = \{(x_l, y_l)\}_{l=1}^N$, we first perform balanced sampling for each expression category. This is because facial expression recognition datasets often exhibit significant class imbalance issues. This can lead to the model overfitting to categories with a large number of samples during training, while potentially underfitting to categories with fewer samples, affecting overall recognition performance. If the model trained in a fully supervised manner itself has such biases, the predicted pseudo-labels will undoubtedly be biased, which will further deteriorate the quality of the model. At the same time, we apply weak data augmentation techniques to improve the generalization ability of the model, and we can obtain the probability distribution $P_l$ predicted by NetworkB:

$$P_l = \text{NetworkB}(\text{WA}(\text{Sample}(x_l)); \theta_B)\qquad(4)$$

where $\text{WA}(\cdot)$ represents the application of weak data augmentation techniques to the data sample. $\text{Sample}(\cdot)$ represents balanced sampling of the data. $\theta_B$ represents the model parameters of NetworkB.

We use $P_l$ and the true expression labels to calculate the cross-entropy loss to train a model without class bias for assigning pseudo-labels to unlabeled data:

$$L_l = \text{CEL}(y_l, P_l)\qquad(5)$$

where $\text{CEL}(\cdot)$ represents the cross-entropy loss function.

For unlabeled data $D_u = \{x_u\}_{u=1}^M$, we first generate two weakly augmented samples $x_{u1}^w$, $x_{u2}^w$ and one strongly augmented sample $x_u^s$ for each sample, then use NetworkA to extract features and infer their respective probability distributions $P_{u1}^w$, $P_{u2}^w$, $P_u^s$:

$$P_{u1}^w = \text{NetworkA}(\text{WA}(x_{u1}); \theta_A)\qquad(6)$$

$$P_{u2}^w = \text{NetworkA}(\text{WA}(x_{u2}); \theta_A)\qquad(7)$$

$$P_u^s = \text{NetworkA}(\text{SA}(x_u); \theta_A)\qquad(8)$$

Next, we obtain the average probability distribution $\bar{P}_u$ of the weakly augmented samples:

$$\widetilde{P}_u = \frac{1}{2}(P_{u1}^w + P_{u2}^w)\qquad(9)$$

If $\text{argmax}(\bar{P}_u)$ is greater than the confidence threshold $\tau$ (where $\tau$ is the confidence threshold for each expression category calculated by the dynamic threshold module for the current training epoch), we obtain the pseudo-label:

$$\hat{y}_u = \text{argmax}(\bar{P}_u)\qquad(10)$$

We use the pseudo-label $\hat{y}_u$ and the strongly augmented sample $P_u^s$ to calculate the consistency loss, encouraging the model prediction to be consistent with the pseudo-label. The consistency loss is calculated as follows:

$$L_u = \text{CEL}(\hat{y}_u, P_u^s)\qquad(11)$$

where $\text{CEL}(\cdot)$ represents the cross-entropy loss function.

For samples where $\text{argmax}(\bar{P}_u)$ is less than the confidence threshold $\tau$, we cannot assign a sufficiently credible pseudo-label to this sample. However, this does not mean that these samples are of poor quality or are noise samples. On the contrary, these samples are likely to be difficult or unseen samples,

## IV. MODULE DESIGN INTRODUCTION

For the original semi-supervised learning framework, our method proposes three improvements to make the traditional semi-supervised learning framework applicable to facial expression recognition tasks. The first is to design a feature extraction network with an added attention mechanism to extract higher quality facial expression features. The second is a dynamic threshold adjustment module, which can fully screen samples that can help facial expression recognition tasks from unlabeled data by dynamically adjusting the threshold for selecting pseudo-labels of different categories. The third is a negative learning module. For the problem of fully utilizing unlabeled data mentioned above, we use a selected negative learning module to help the model mine information that can help us perform facial expression recognition from samples with low confidence.

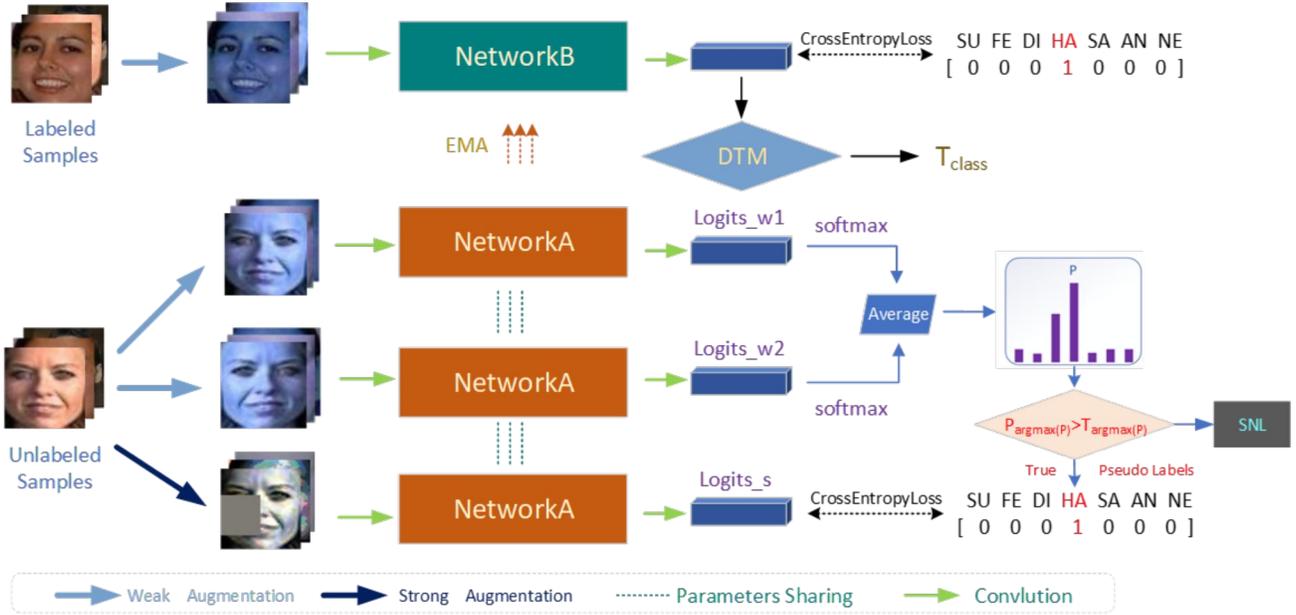

Fig. 2. Semi-supervised recognition framework based on dynamic threshold and selective negative learning. We adopted a semi-supervised approach, uniformly sampling the number of samples for each emotion category to reduce model bias caused by the imbalanced distribution of sample quantities. Our designed framework is capable of assigning pseudo-labels to unlabeled data, helping the model learn facial expression features. Considering the varying difficulty in recognizing each category, we set a confidence threshold for each emotion category and gradually adjust this threshold as the training progresses to fully leverage the unlabeled data. To maximize the value of unlabeled data, we specifically designed a selective negative learning module, aimed at fully mining the information from samples with low confidence by extracting beneficial information for expression recognition from complementary labels.

### A. Feature Extraction Network

In the field of facial expression recognition, many methods [3], [4] use attention mechanisms to enhance the model's focus on local areas, thereby improving the accuracy of facial expression recognition. However, excessive attention mechanisms may cause the model to pay redundant attention to some facial areas, causing the model to focus only on very few local information, thereby ignoring more facial area features, especially some areas that have the potential to assist facial expression recognition tasks. Especially when some key areas are partially occluded or some facial areas have large posture changes, the model may not have enough generalization ability, leading to inaccurate recognition.

Considering this problem, we hope that the feature extraction network can extract more diverse and more global features. Similar to the dropout operation of convolutional neural networks, we designed a feature-level dropout module. After extracting N feature maps, we randomly select a feature map to drop with a probability of p. The purpose of this is to prevent the model from only focusing on local features, forcing the model to lose some attention information, and instead focus on other facial areas for facial expression recognition.

The schematic diagram of the feature extraction network is shown in Figure 3.

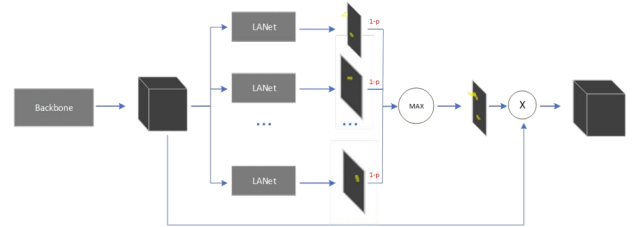

Fig. 3. We aim for the feature extraction network to capture more diverse and global features. Similar to the dropout operation in convolutional neural networks, we designed a feature-level dropout module. After extracting N feature maps using LANet, we randomly discard one feature map with a probability of p. The purpose of this approach is to prevent the model from focusing solely on local features, forcing it to lose some attention information and instead enhance its focus on other facial regions for facial expression recognition.

We process the input feature map with multiple Local Attention Networks (LANet). For each LANet, the input feature map is reduced in channel number and feature extraction through two $1 \times 1$ convolution layers. The first convolution layer reduces the channel number from $c$ to $c/r$, and the second convolution layer further reduces the channel number to 1. This process can be simplified as:

$$F_{score} = LANet((F_{input})) \qquad (12)$$

where $F_{input} \in R^{C \times H \times W}$ is the input feature map, and $F_{score} \in R^{1 \times h \times w}$ is the attention weight map.

---

**Algorithm 1:** Drop process of attention map in feature extraction network

---

**Input:** inputs ($F_{score}$)
**Output:** outputs ($F_{atten}$)

1 **Function** Main(*inputs*):
    # Get the feature map output by the last layer of LANet
2   **if** *inputs is a tuple* **then**
3     |  inputs ← inputs[-1]
4   **end**
    # Get batch size
5   bs ← size of first dimension of inputs
    # Randomly select a feature map for each group of features in the batch
6   index ← random integers in [0, LANet_num] for bs times
7   **for** *i ← 0 to bs-1* **do**
      # Discard the selected feature map with a probability of p
8     **if** *random number in [0, 1) ¿ (1 - p)* **then**
9       |  inputs[i, index[i], . . .] ← 0
10     **end**
11   **end**
    # After discarding some features, apply the max operation to the remaining features
12   outputs ← max(inputs, 1, keepdim=True)[0]
13   **return** outputs

---

In order for the model to refer to more facial information, similar to the Dropout operation. For the output of each LANet, we randomly select one of the feature maps and discard this feature map with a probability of p. Then we take the max operation bit by bit on the remaining feature maps to obtain the maximum attention weight map $F_{atten}$. The detailed process can refer to Algorithm 1.

Finally, we calculate the Hadamard product of the attention map $F_{atten}$ output by the local area perception module and the output $F_{input}$ of the backbone network to obtain the local attention-enhanced facial expression feature $F_{exp}$, that is:

$$F_{exp} = F_{input} \odot F_{atten},  \tag{13}$$

where $\odot$ represents the Hadamard product.

### B. Dynamic Threshold Adjustment (DTA)

To enhance the robustness of the model, we train a stable facial expression recognition model using the Exponential Moving Average (EMA) technique with a decay rate of 0.999, as shown in Figure 2. We use this exponentially weighted averaged model, NetworkB, to extract features and probability distributions of the labeled data $(x_l, y_l)$:

$$\tilde{P}_l = \text{NetworkB}(x_l; \theta_B)  \tag{14}$$

Then, we calculate the average confidence score of all correctly predicted samples for each category in the labeled data:

$$\tau_c = \frac{1}{N_c} \sum_{i=1}^{N_c} p_i^c  \tag{15}$$

where $N_c$ represents the total number of correctly predicted samples for class c, $p_i^c$ is the confidence score of the ith sample for the correct class c, and $\tau_c$ is the threshold for class c. In this way, we obtain a confidence threshold for each emotional category in each round.

As the number of training epochs increases, the model's ability to distinguish training data significantly improves. To prevent the threshold from growing too quickly, we perform a weighted average on the threshold to prevent sudden changes. Therefore, the final confidence threshold $\tau_c^t$ for class i in each epoch is:

$$\tau_c^t = \mu \tau_c^{t-1} + (1 - \mu) \tau_c  \tag{16}$$

where $\mu$ is a hyperparameter.

### C. Selective Negative Learning (SNL)

For samples where $\text{argmax}(P^-_u)$ is less than the confidence threshold $\tau$, intuitively, we consider the predicted category of the sample to be unreliable and cannot directly use it to improve the model's ability to recognize expressions. If the confidence is low, the pseudo-label of the sample is likely to be incorrect, and forcing consistency learning with strongly augmented samples can lead to model confusion between the pseudo-label and the true label of the sample, affecting the model's judgment of category boundaries. To make full use of these samples, we perform selective negative learning to mine useful information from samples with low confidence.

For those unreliable samples, although predicting their expression categories is challenging, there should be some categories with sufficiently low probability scores. Therefore, the model can easily determine which categories these unreliable samples do not belong to, which are called complementary labels. To obtain complementary labels as comprehensively as possible, we maintain a complementary label library and update it progressively. Specifically, for an unreliable sample's prediction, we find the minimum probability value in each category. If the minimum probability value is less than the threshold $\delta$, we add its corresponding category to the complementary label library and accordingly obtain the complementary label $\bar{y}^c$ as follows:

$$\bar{y}^c = \begin{cases} 1, & \text{if } min(\tilde{P}_u) \leq \delta \text{ and } c = \text{argmin}(\tilde{P}_u) \\ 0, & \text{otherwise} \end{cases}  \tag{17}$$

where $\bar{y}^c$ represents the value of the one-hot label for the expression category c. $\delta$ is a positive constant to ensure enough confidence in assigning one-hot labels in complementary labels. The complementary label $\bar{y}^c$ is then used to train the model through negative learning:

$$L_{NL} = -\sum_{c \in C} \bar{y}^c (1 - \tilde{P}_u^c),  \tag{18}$$

where $P_u^{-c}$ represents the predicted probability for category c. The above process will be iterated until the remaining probabilities are all above a specific threshold $\delta$. It is worth noting that categories in previous complementary labels will not be involved in the probability prediction of the next iteration.

### D. Loss Function

This paper involves supervised learning with labeled data, supervised learning with pseudo-labeled data, and negative learning tasks with low-confidence samples. Therefore, the loss function of the semi-supervised learning framework proposed is:

$$L_{final} = L_l + \lambda_1 L_u + \lambda_2 L_{NL} \quad (19)$$

## V. EXPERIMENTS

### A. Datasets

RAF-DB [5] is a publicly available facial expression recognition dataset, designed specifically for capturing and recognizing emotional expressions in the real world. The dataset contains $15,339$ annotated high-quality images collected from the Internet, covering a variety of facial expressions collected from open environments. The expressions in the images are divided into seven basic emotional categories: happiness, sadness, surprise, fear, anger, disgust, and neutral, providing rich resources for facial expression recognition research. Each image is annotated by independent annotators for emotional categories to ensure the accuracy and consistency of the annotations.

AffectNet [6] dataset is a larger and more diverse facial expression database, containing over a million images with facial expressions. The images are collected from the Internet and annotated by experts for expression categories. The dataset covers various lighting conditions, postures, ages, genders, and races. Like the RAF-DB dataset, we used the data of the seven annotated categories: happiness, sadness, surprise, fear, anger, disgust, and neutral.

These two datasets are commonly used public datasets in the field of facial expression recognition, and evaluating them is beneficial for our comparison with other methods. These two datasets have two main difficulties. The first is the class imbalance situation. As can be seen from Table I, both datasets have a large gap in the number of samples in each category. Compared with RAF-DB, the class imbalance phenomenon of AffectNet is more serious, the number of samples in the "happiness" category is more than 30 times the number of samples in the "disgust" category. The second difficulty is that both datasets are data collected in an open environment. The images themselves are affected by the external environment's lighting, background, and facial posture, which poses a significant challenge to the model's learning ability.

### B. Data Preprocessing

Our framework conducted semi-supervised training experiments on the RAF-DB and AffectNet datasets. All facial expression images were first resized to $256 \times 256$, and then

#### TABLE I
#### NUMBER OF SAMPLES IN EACH EXPRESSION CATEGORY IN RAF-DB AND AFFECTNET DATASETS

|           | RAF-DB | AffectNet |
|-----------|--------|-----------|
| Happiness | 5957   | 134915    |
| Sadness   | 2460   | 25959     |
| Surprise  | 1619   | 14590     |
| Fear      | 355    | 6878      |
| Anger     | 867    | 25382     |
| Disgust   | 877    | 4303      |
| Neutral   | 3204   | 75374     |

during training, they were randomly cropped to $224 \times 224$ images and input into the feature extraction network. The weak data augmentation method used in this paper is random cropping and random horizontal flipping, and the strong data augmentation method is RandAugment [32]. Referring to the RandAugment method, we constructed 14 data augmentation methods as shown in Table II. We use the data augmentation method RandAugment(3,5) to indicate that 3 of the 14 data augmentation methods are selected, and the intensity is 5.

#### TABLE II
#### 14 DATA AUGMENTATION METHODS USED IN RANDAUGMENT

| Data Augmentation Method | Data Augmentation Method |
|--------------------------|--------------------------|
| Rotate                   | Sharpness                |
| Shear-x                  | Shear-y                  |
| Translate-x              | Translate-y              |
| Identity                 | Contrast                 |
| Color                    | Brightness               |
| Equalize                 | Solarize                 |
| Posterize                | AutoContrast             |

To be consistent with other papers, we set the RAF-DB dataset to have 100 labels, 400 labels, 2000 labels, and 4000 labels. That is, only the expression category labels of these data are used. For the AffectNet dataset, due to the large scale of the dataset and for comparison with mainstream methods, we chose the experimental settings of 2000 labels and 10000 labels. Due to the small amount of data and uneven data distribution, for comparison with existing methods, for these experimental settings, we listed the number of samples of each expression category in the RAF-DB dataset in Table III. The AffectNet dataset is relatively large in scale, so each category of data can be sampled evenly, and the number of images in each category does not exceed one.

#### TABLE III
#### NUMBER OF SAMPLES IN EACH EXPRESSION CATEGORY UNDER SEMI-SUPERVISED SETTINGS IN RAF-DB DATASET

|                            | 100labels | 400labels | 2000labels | 4000labels |
|----------------------------|-----------|-----------|------------|------------|
| Fear                       | 10        | 40        | 200        | 250        |
| Other expression categories| 15        | 60        | 300        | 625        |

### C. Evaluation Metrics

We use accuracy to evaluate the performance of our model. Accuracy is a metric used to evaluate the performance of classification models, defined as the number of correctly predicted samples divided by the total number of samples. When comparing with other methods, to align with other papers, we set 5 random seeds for the experiment and take the average accuracy ± standard deviation for evaluation.

### D. Experimental Settings

To make a fair comparison with other semi-supervised methods, we chose the resnet18 model pre-trained on the MS-Celeb-1M [33] face recognition dataset as our backbone network. The neural network optimizer uses Adam, the learning rate is set to 0.0005, the batch size is set to 128, and a total of 30 rounds are trained. The loss function weighting coefficients $\lambda_1$ is set to 0.5, and $\lambda_2$ is set to 0.1. The hardware equipment for the experiment is NVIDIA V100S.

### E. Experimental Results

*1) Validation Results on RAF-DB Dataset:* We use the experimental results of RAF-DB labeled data as the baseline experimental results. To make a fair comparison with other papers, the fully supervised recognition results on RAF-DB come from DLP-CNN [5]. The data in the table shows that our scheme has achieved good results. Compared with other semi-supervised recognition methods, our method has a significant improvement in 4 experimental settings. Among them, our method has obvious advantages in the experimental settings of 100 labels and 4000 labels. For the experimental settings corresponding to 400 labels and 2000 labels, our method is still 1% to 2% behind the current facial expression recognition semi-supervised methods. The data in the table shows that our method can surpass the performance of fully supervised learning without using all the data.

TABLE IV

COMPARISON OF EXPERIMENTAL RESULTS ON RAF-DB DATASET WITH CURRENT MAINSTREAM SEMI-SUPERVISED LEARNING METHODS

| Method | RAF-DB | | | |
|---|---|---|---|---|
| | 100 labels | 400 labels | 2000 labels | 4000labels |
| Baseline | 52.43±2.24 | 67.75±0.95 | 78.91±0.43 | 81.90±0.48 |
| Pseudo-Labeling [24] | 54.96±4.24 | 69.99±1.81 | 79.18±0.27 | 82.88±0.49 |
| MixMatch [34] | 54.57±4.16 | 73.14±1.40 | 79.63±0.91 | 83.57±0.49 |
| UDA [35] | 58.15±1.54 | 72.39±1.64 | 81.16±0.54 | 83.56±0.82 |
| ReMixMatch [36] | 58.83±2.34 | 73.34±1.82 | 79.66±0.66 | 83.51±0.18 |
| MarginMix [37] | 58.91±1.78 | 73.31±1.64 | 80.22±0.76 | 83.47±0.28 |
| FixMatch [29] | 60.67±2.25 | 73.36±1.59 | 81.27±0.27 | 83.31±0.33 |
| Ada-CM [38] | 62.36±1.10 | 74.44±1.53 | 82.05±0.22 | 84.42±0.49 |
| Ours | **64.06±1.86** | **73.27±1.36** | **80.24±0.13** | **84.45±0.36** |
| Fully Supervised | 84.13 | | | |

*2) Validation Results on AffectNet Dataset:* The RAF-DB dataset is relatively small in scale. We conducted experiments on the AffectNet dataset, which is 20 times the size of the RAF-DB dataset. We also use the experimental results of AffectNet labeled data as the baseline experimental results. To make a fair comparison with other papers, the fully supervised recognition results on AffectNet come from SAN [39]. The experimental results show that our method has achieved very good results on the AffectNet dataset. Our experimental method has an average accuracy of 2.2% higher than the second place in the 2000 labels experimental setting, and an average accuracy of 1.01% higher than the second place in the 10000 labels experimental setting, which fully demonstrates the superiority of our scheme. Our method can surpass the experimental results of full supervision when only using 2000 labels, and the training set has 280,000 pictures and corresponding expression labels, which fully proves the effectiveness of our semi-supervised scheme.

TABLE V

COMPARISON OF EXPERIMENTAL RESULTS ON AffectNet DATASET WITH CURRENT MAINSTREAM SEMI-SUPERVISED LEARNING METHODS

| Method | Affectnet | |
|---|---|---|
| | 2000 labels | 10000 labels |
| Baseline | 47.52±0.75 | 53.18±0.68 |
| Pseudo-Labeling [24] | 48.78±0.67 | 53.82±1.29 |
| MixMatch [34] | 49.63±0.49 | 53.49±0.47 |
| UDA [35] | 50.42±0.45 | 56.49±0.27 |
| ReMixMatch [36] | 50.38±0.63 | 55.81±0.34 |
| MarginMix [37] | 50.58±0.42 | 56.41±0.28 |
| FixMatch [29] | 50.79±0.37 | 56.50±0.43 |
| Ada-CM [38] | 51.22±0.29 | 57.42±0.43 |
| LION [40] | 52.71±0.21 | 59.11±0.38 |
| Ours | **54.91±0.25** | **60.12±0.23** |
| Fully Supervised | 52.97 | |

*3) Module Ablation Experimental Results:* We conducted experiments on various modules proposed under the experimental settings of 100 labels and 4000 labels on the RAF-DB dataset. When only using labeled data, our method achieved average accuracies of 58.63% and 82.90% respectively. The data in the table shows that our unlabeled data, selective negative sampling, feature enhancement network, and dynamic threshold methods all produced positive gains for facial expression recognition results. Among them, the improvement of the feature enhancement network is particularly noticeable. Without using feature enhancement, the highest experimental result of 100 labels is only 63.05%, but after using the feature enhancement network, it has increased by at least 1.53%. In the 4000 labels experimental setting, without using feature enhancement, the highest accuracy can reach 83.78%, while after using it, the lowest reached 84.22%, which clearly shows the significant improvement brought by the feature enhancement network.

TABLE VI
MODULE ABLATION EXPERIMENT

| $L_l$ | $L_u$ | $L_{NL}$ | Feature Enhancement | Dynamic Threshold | 100 labels | 4000 labels |
|---|---|---|---|---|---|---|
| ✓ | | | | | 58.63 | 82.90 |
| ✓ | ✓ | | | | 61.64 | 83.55 |
| ✓ | ✓ | ✓ | | | 62.17 | 83.78 |
| ✓ | ✓ | ✓ | ✓ | | 64.58 | 84.22 |
| ✓ | ✓ | ✓ | | ✓ | 63.05 | 83.97 |
| ✓ | ✓ | ✓ | ✓ | ✓ | **65.10** | **84.38** |

*4) Feature Enhancement Network:* We also considered the impact of the number of LANet branches in the feature enhancement network on the experimental results. We set the number of LANet branches $N = 0, 2, 4, 6, 8, 10, 12$, and verified it under the experimental setting of 4000 labels on the RAF-DB dataset. The experimental results are shown in Figure 4. After experimental verification, the effect is best when $N = 6$. When the number of LANet branches is small, the attention mechanism is not saturated and can learn the features of new patterns. When the number of LANet branches reaches a certain level, increasing the number of branches will not help the model notice more discriminative features.

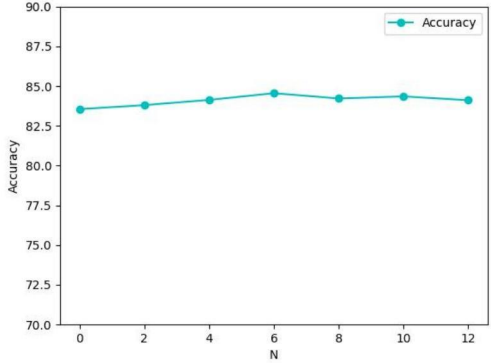

Fig. 4. Impact of the Number of LANet Branches.

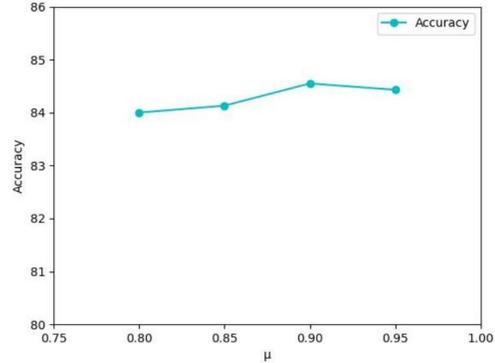

Fig. 5. Impact of Weighting Coefficient in Dynamic Threshold Module.

TABLE VII
Impact of Probability p in Feature Map Random Dropout Strategy

| Method | Affectnet | |
|---|---|---|
| | 2000 labels | 10000 labels |
| 0 | 54.45 | 60.17 |
| 0.3 | 55.17 | **60.40** |
| 0.5 | **55.39** | 60.39 |
| 0.7 | 54.60 | 60.19 |

The random dropout strategy of the feature map in the feature enhancement network is also important, and the most critical of this strategy is the setting of probability p. We designed an ablation experiment to verify the impact of the probability value. The experiment in Table VII is the result on the AffectNet dataset. The experimental results show that using random dropout strategy has a positive gain for the experiment results. In the 2000 labels setting, the best result is obtained when p is 0.5. In the 10000 labels setting, the best result is obtained when p is 0.3, but the difference from when p is 0.5 is very small, only 0.01%, so this article sets p to 0.5 for the experiment.

*5) Dynamic Threshold Adjustment:* In the dynamic threshold module, the threshold of each round is the exponential weighted average of the current round and the previous round. We considered the impact of the weighting coefficient $\mu$ on the experimental results. We set $\mu = 0.8, 0.85, 0.9, 0.95$, and verified it under the experimental setting of 4000 labels on the RAF-DB dataset. The experimental results are shown in Figure 5. From the data in the figure, when the weighting coefficient $\mu$ is set to 0.9, the effect is optimal.

### F. Cross-Dataset Evaluation

We also conducted cross-dataset experiments, using the Affwild2 [41] dataset as the labeled data source. This database is in total consists of 548 videos of around 2.7M frames that are annotated in terms of the 6 basic expressions (i.e., anger, disgust, fear, happiness, sadness, surprise), plus the neutral state and a category 'other' that denotes expressions other than the 6 basic ones. Due to the substantial scale of this dataset, we sampled 8000 images per class for experimentation. For the unlabeled data, considering the dataset's size, we utilized samples from the Affectnet dataset. Additionally, recognizing the significantly larger scale of face recognition datasets compared to facial expression datasets, we explored using a Face Recognition dataset as unlabeled data in our experiments.

Specifically, for the Affwild2 dataset, we utilized 64,000 images for training and 280,000 images for validation. Affectnet contributed an additional 280,000 images to our dataset. Regarding the face recognition data, we randomly sampled 500,000 images from the MS-Celeb-1M dataset to serve as unlabeled data for training. Moreover, we employed the EfficientNet-B7 [42] as the backbone for our model. To better showcase the performance of our model, we also incorporated the macro-F1 score as an evaluation metric. This comprehensive approach involving multiple datasets and a diverse set of images aims to ensure the robustness and effectiveness of our proposed model across different scenarios.

TABLE VIII
Cross-dataset experimental results.

| Aff-wild2 | Affectnet | MS-Celeb-1M | Accuracy | F1 score |
|---|---|---|---|---|
| ✓ | | | 48.54 | 34.80 |
| ✓ | ✓ | | 50.90 | 37.26 |
| ✓ | | ✓ | 49.65 | 38.68 |
| ✓ | ✓ | ✓ | 52.72 | 40.45 |

## VI. Conclusion

The application prospects of facial expression recognition in human-computer interaction, public safety, mental health, and leisure and entertainment are very broad. It can be said to be an indispensable part of future artificial intelligence technology. This manuscript first analyzes the technical difficulties existing in the current facial expression recognition task, mainly including two major problems of limited data scale and unbalanced dataset classes. Then, we introduce the solution we propose. In response to the problem of limited data scale, we hope to achieve better results with less data. Therefore, this manuscript introduces semi-supervised learning methods. In order to make the semi-supervised learning framework and facial expression recognition tasks more compatible, we designed a dynamic threshold adjustment module to optimize the way to screen pseudo-labels for facial expression recognition tasks. At the same time, in the process of screening pseudo-labels, we designed a selective negative learning module to fully tap the information of those low-confidence samples. In response to the problem of unbalanced dataset classes, we use a joint optimization method of labeled data and unlabeled data, use balanced sampling of labeled data to train the model, so that the model can reduce the bias between categories when assigning pseudo labels, thereby suppressing the problem that the model bias deepens continuously. In addition, we also involve a feature extraction network, which successfully enhances the model's feature extraction ability through local feature enhancement and feature map random dropout strategies.

Finally, through experiments on the RAF-DB and AffectNet datasets, and comparisons with current mainstream methods, this manuscript verifies the effectiveness of our semi-supervised facial expression recognition framework.

## ACKNOWLEDGMENTS

This work was supported by the Natural Science Foundation of China (62276242, 62372153), National Aviation Science Foundation (2022Z071078001), CAAI-Huawei Mind-Spore Open Fund (CAAIXSJLJJ-2022-001A), Anhui Province Key Research and Development Program (202104a05020007), Dreams Foundation of Jianghuai Advance Technology Center (2023-ZM01Z001), USTC-IAT Application Sci. & Tech. Achievement Cultivation Program (JL06521001Y).